\documentclass[]{bytedance_seed}

\usepackage[toc,page,header]{appendix}
\usepackage{xcolor}

\tcbset{
  cleanScenario/.style={
    colback=white,              
    colframe=blue!40!white,     
    boxrule=1.2pt,              
    arc=1mm,                    
    left=2mm, right=2mm, top=1mm, bottom=1mm,
    enhanced, sharp corners     
  }
}

\usepackage{minitoc}
\usepackage{color}
\usepackage{graphicx}
\usepackage{subcaption}
\usepackage{caption}
\usepackage{float}
\usepackage{tabularx}
\usepackage{booktabs}
\usepackage{pifont}
\usepackage{amssymb}
\usepackage{makecell}
\usepackage{tikz}
\usepackage{xltabular} 
\usepackage{ragged2e}

\newcolumntype{L}[1]{>{\RaggedRight\arraybackslash\hsize=#1\hsize}X}

\usepackage{amsmath}
\usepackage{enumitem}

\usepackage[most]{tcolorbox}

\usepackage{titlesec}

\definecolor{ModelTitleBG}{RGB}{70, 130, 180}     
\definecolor{ModelFrame}{RGB}{176, 196, 222}   
\definecolor{ModelContentBG}{rgb}{0.98, 0.98, 1.0} 
\definecolor{SectionTitleColor}{RGB}{47, 79, 79} 

\newtcolorbox{modelbox}[1]{
    enhanced, 
    breakable, 
    title=#1, 
    attach boxed title to top left={yshift=-2mm, xshift=3mm}, 
    colback=ModelContentBG, 
    colframe=ModelFrame, 
    colbacktitle=ModelTitleBG, 
    coltitle=white, 
    fonttitle=\bfseries\large, 
    rounded corners, 
    drop fuzzy shadow, 
    boxrule=1pt, 
    top=8mm, 
    left=4mm,
    right=4mm,
    bottom=4mm,
    before upper={\small},
}

\title{ FutureX-Pro: Extending Future Prediction to High-Value Vertical Domains }

\affiliation[]{ByteDance Seed}
\affiliation[]{Hong Kong University of Science and Technology}
\affiliation[]{Georgia Institute of Technology}
\affiliation[]{Stanford University}
\affiliation[]{Princeton University}

\contribution{Full author list in Contributions}
\vspace{-0.15in}

\abstract{
Building upon FutureX, which established a live benchmark for general-purpose future prediction, this report introduces \textbf{FutureX-Pro}, including \textbf{FutureX-Finance}, \textbf{FutureX-Retail}, \textbf{FutureX-PublicHealth}, \textbf{FutureX-NaturalDisaster}, and \textbf{FutureX-Search}.
These together form a specialized framework extending agentic future prediction to high-value vertical domains. 
While generalist agents demonstrate proficiency in open-domain search, their reliability in capital-intensive and safety-critical sectors remains under-explored. FutureX-Pro targets four economically and socially pivotal verticals: Finance, Retail, Public Health, and Natural Disaster. We benchmark agentic Large Language Models (LLMs) on entry-level yet foundational prediction tasks—ranging from forecasting market indicators and supply chain demands to tracking epidemic trends and natural disasters. By adapting the contamination-free, live-evaluation pipeline of FutureX, we assess whether current State-of-the-Art (SOTA) agentic LLMs possess the domain grounding necessary for industrial deployment. Our findings reveal the performance gap between generalist reasoning and the precision required for high-value vertical applications. 
}

\vspace{-0.3in}
\correspondence{Jiashuo Liu \email{liujiashuo77@gmail.com}, Wenhao Huang \email{huang.wenhao@bytedance.com}}

\checkdata[Project Page]{\url{https://futurex-ai.github.io/}}

\begin{document}
\maketitle

\section{Introduction}

The evolution of LLM agents has enabled systems that can gather information and reason about the future. 
Our previous work, FutureX~\cite{zeng2025futurex}, evaluates these capabilities across a broad spectrum of general topics and has sparked extensive discourse with \emph{over 100 million impressions}. 
Yet, the true test of agentic intelligence lies beyond generalist tasks—it resides in domains that dictate economic and societal stability. 
\textbf{FutureX-Pro} addresses this critical gap by extending future prediction to high-value vertical domains, explicitly shifting the evaluation focus from general breadth to high-stakes depth.

FutureX-Pro focuses on four verticals chosen not just for their complexity, but for their profound impact on global systems:
\begin{itemize}
    \item \emph{Finance (Market Efficiency)}: Financial markets represent the pulse of the global economy. The ability to accurately predict entry-level indicators, such as corporate revenue or macroeconomic indices, is fundamental to understanding economic health. This domain demands rigorous quantitative reasoning and the ability to filter noise from high-frequency market signals.
    \item \emph{Retail (Resource Optimization)}: Retail forecasting drives the efficiency of global supply chains. Predicting product demand and sales trends is crucial for reducing inventory waste and optimizing resource distribution. This domain tests an agent’s ability to interpret consumer behavior and dynamic market trends.
    \item \emph{Public Health (Societal Resilience)}: In the wake of global health challenges, the capacity to track infectious diseases is a cornerstone of societal resilience. Accurate prediction of epidemic metrics supports early warning systems and public health preparedness. Evaluating agents here validates their potential reliability in biosurveillance contexts.
    \item \emph{Natural Disaster (Environmental Safety)}: As climate volatility increases, predicting the occurance and impact of environmental hazards (e.g., typhoons, floods) becomes a critical safety imperative. Success in this domain relies on interpreting meteorological data to forecast risks, serving as a proxy for an agent's utility in high-stakes safety scenarios.
\end{itemize}

While recent agentic LLMs excel at synthesizing diverse, open-web information, high-value verticals impose distinct challenges regarding data heterogeneity and source authority. A generalist agent proficient in summarizing political news may fail to correctly interpret a structured financial disclosure, a meteorological chart, or an epidemiological table. FutureX-Pro investigates this gap, evaluating whether agents can move beyond generic web search to effectively ground their reasoning in authoritative, domain-specific data sources.

FutureX-Pro leverages the ``contamination-impossible by design'' architecture of FutureX, ensuring robust evaluation by focusing on verifiable future events. 
The benchmark assesses foundational competence through tasks that simulate the core responsibilities of a junior analyst:
\begin{itemize}
    \item \emph{Finance}: Analyzing real-time reporting to forecast short-term financial performance.
    \item \emph{Retail}: Synthesizing e-commerce data to predict sales rankings and trends.
    \item \emph{Public Health}: Interpreting official bulletins to project near-term disease metrics.
    \item \emph{Natural Disaster}: Processing meteorological data to forecast hazard impacts.
\end{itemize}

This report presents the design and construction of the domain-specific datasets and a comparative analysis of leading agentic LLMs. 
We aim to provide a clear assessment of how current AI systems perform when tasked with reasoning about the high-value domains that shape our world.

Complementing our predictive benchmarks, we also design \textbf{FutureX-Search}, which leverages historical data from FutureX and FutureX-Pro to transform predictive tasks into pure search and retrieval challenges for resolved events. 
The remainder of this report provides a detailed breakdown of the four vertical domains and the FutureX-Search benchmark\footnote{The scope of this report is limited to the introduction of FutureX-Pro and FutureX-Search. As for the detailed construction process, please refer to FutureX~\cite[Section 3.2]{zeng2025futurex}.}.

\textbf{Crypto Domain.} For evaluations in the cryptocurrency domain, please refer to our dedicated benchmark, CryptoBench~\cite{guo2025cryptobench}.

\textbf{Pipeline Implementation.} For details on the construction pipeline of FutureX-Pro, please refer to the methodology in FutureX~\cite{zeng2025futurex}.

\textbf{Live Updates.} The new vertical tasks will soon be incorporated into our weekly FutureX competition at \url{https://huggingface.co/datasets/futurex-ai/Futurex-Online}.

\section{FutureX-Finance}
\label{sec:finance}

Financial forecasting represents the pinnacle of complex adaptive systems, where agents must navigate volatility, decipher market sentiment, and process quantitative reports~\citep{hu2025finsearchcomp, zhu2025findeepresearch, dong2025finch}. \textbf{FutureX-Finance} evaluates an agent's capability to function as a quantitative analyst, requiring precise numerical predictions of stock market indicators across global exchanges.

\subsection{Data Construction and Market Structure}

To ensure a comprehensive evaluation of global economic understanding, we curate a dataset encompassing major equities from the world's two largest economies. The dataset tracks a diverse portfolio of \emph{150 distinct entities}, selected to represent the structural core of each market.

\textbf{Market Coverage.}
\begin{itemize}
    \item \emph{US Market (NASDAQ/NYSE):} We select \emph{100 constituent companies} from the NASDAQ-100 and S\&P 500 indices. This selection includes global tech giants like \textit{NVIDIA (NVDA)} and \textit{Microsoft (MSFT)}, as well as biotech leaders like \textit{Vertex (VRTX)}.
    \item \emph{China A-Share Market (SSE/SZSE):} We select \emph{50 top-tier firms} representing the backbone of the Chinese economy, such as \textit{Kweichow Moutai (600519.SH)} and \textit{CATL (300750.SZ)}.
\end{itemize}

\begin{figure}[h]
    \centering
    \includegraphics[width=\linewidth]{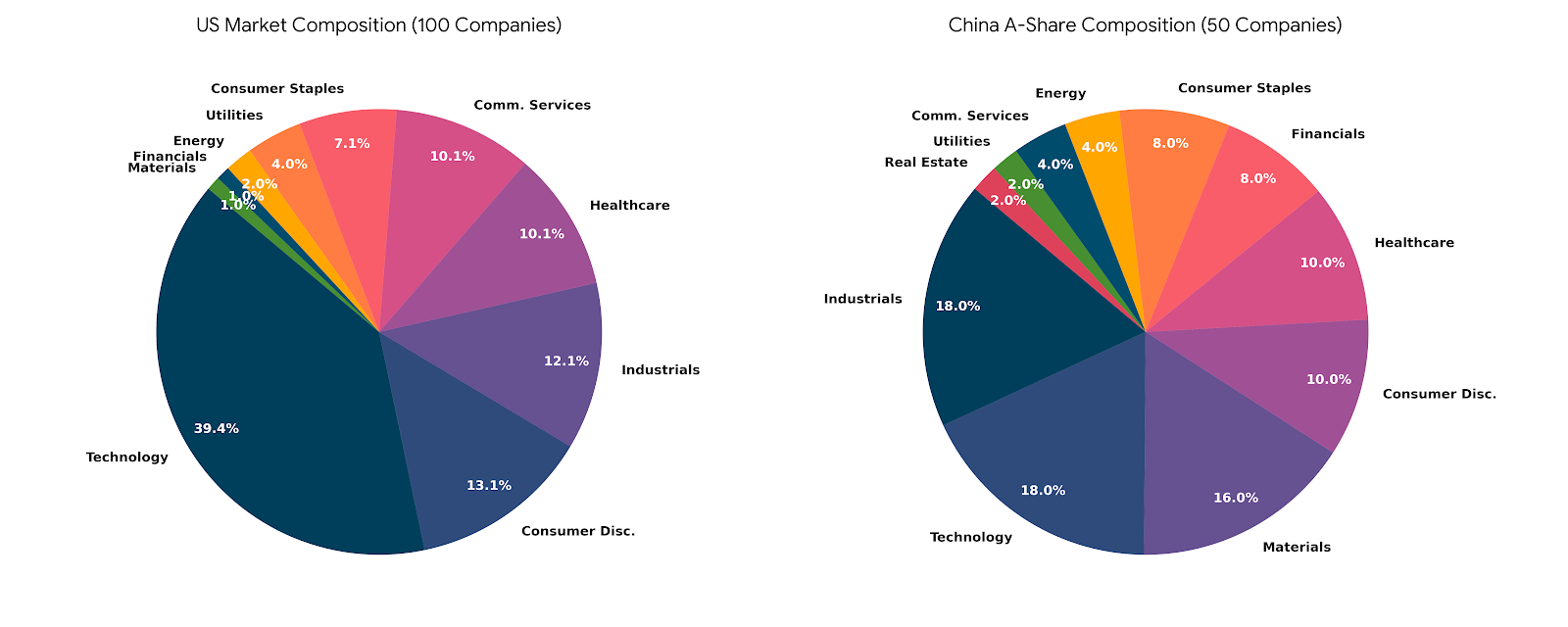}
    \caption{Sector distribution of the 150 tracked entities. The US subset emphasizes Technology, while the China subset highlights Industrials and Materials.}
    \label{fig:finance_sector}
\end{figure}

\textbf{Sectoral Composition Analysis.}
As shown in Figure~\ref{fig:finance_sector}, the dataset mirrors the distinct economic characteristics of the two nations. 
\begin{itemize}
    \item The \emph{US selection} is heavily weighted towards \emph{Technology} and \emph{Communication Services}, reflecting the innovation-driven nature of the US equity market.
    \item The \emph{China selection} exhibits a more balanced distribution with significant representation in \emph{Industrials}, \emph{Materials}, and \emph{Consumer Staples}, capturing the manufacturing and consumption fundamentals of the Chinese economy.
\end{itemize}
This dual-market structure tests the agent's ability to adapt its forecasting logic across different regulatory environments and economic drivers.

\textbf{Task Formulation.}\quad We define three distinct predictive tasks that simulate the workflow of a trader assessing short-term market movements. 

\paragraph{Task Type 1: Spot Prediction.}
The agent must forecast the exact value of a specific term for a single target trading day.
\begin{quote}
    \textit{``What will be the \{Closing Price\} of \{Tesla, Inc. (ID: TSLA)\} on \{2025-12-25\}?''}
\end{quote}

\paragraph{Task Type 2: Window Extremum Prediction.}
The agent must identify the maximum absolute value of a term across a future window of $N$ trading days. This tests the ability to model volatility ranges.
\begin{quote}
    \textit{``What will be the \{highest price\} of \{China Mobile (ID: 600941.SH)\} across \{5\} trading days starting from ...?''}
\end{quote}

\paragraph{Task Type 3: Directional Momentum Prediction.}
The agent focuses on price changes, for example, predicting the highest positive change in a specified window.
\begin{quote}
    \textit{``What will be the \{highest Positive Change\} of \{NVIDIA (ID: NVDA)\} across \{3\} trading days...?''}
\end{quote}

\subsection{Evaluation Metrics}

Financial markets demand extreme precision; a small percentage error can lead to significant arbitrage loss. Therefore, we use a \emph{High-Sensitivity Error Metric}.

Let $\hat{y}$ be the predicted value and $y$ be the actual market value. The score $S$ is calculated using a steep linear penalty function:

\begin{equation}
    S = \max \left( 0, 1 - 20 \times \frac{|\hat{y} - y|}{|y|} \right)
\end{equation}

\paragraph{Interpretation.}
The coefficient of ``20'' imposes a strict tolerance window. 
\begin{itemize}
    \item A relative error of 1\% ($0.01 \times 20 = 0.2$) results in a score of 0.8.
    \item A relative error of 5\% ($0.05 \times 20 = 1.0$) results in a score of 0.0.
\end{itemize}
Any prediction deviating by more than 5\% from the ground truth receives zero points. This rigorous standard ensures that high scores reflect genuine market insight rather than lucky approximations.

\subsection{Results}

\begin{figure}[htbp]
    \centering
    \includegraphics[width=0.7\linewidth]{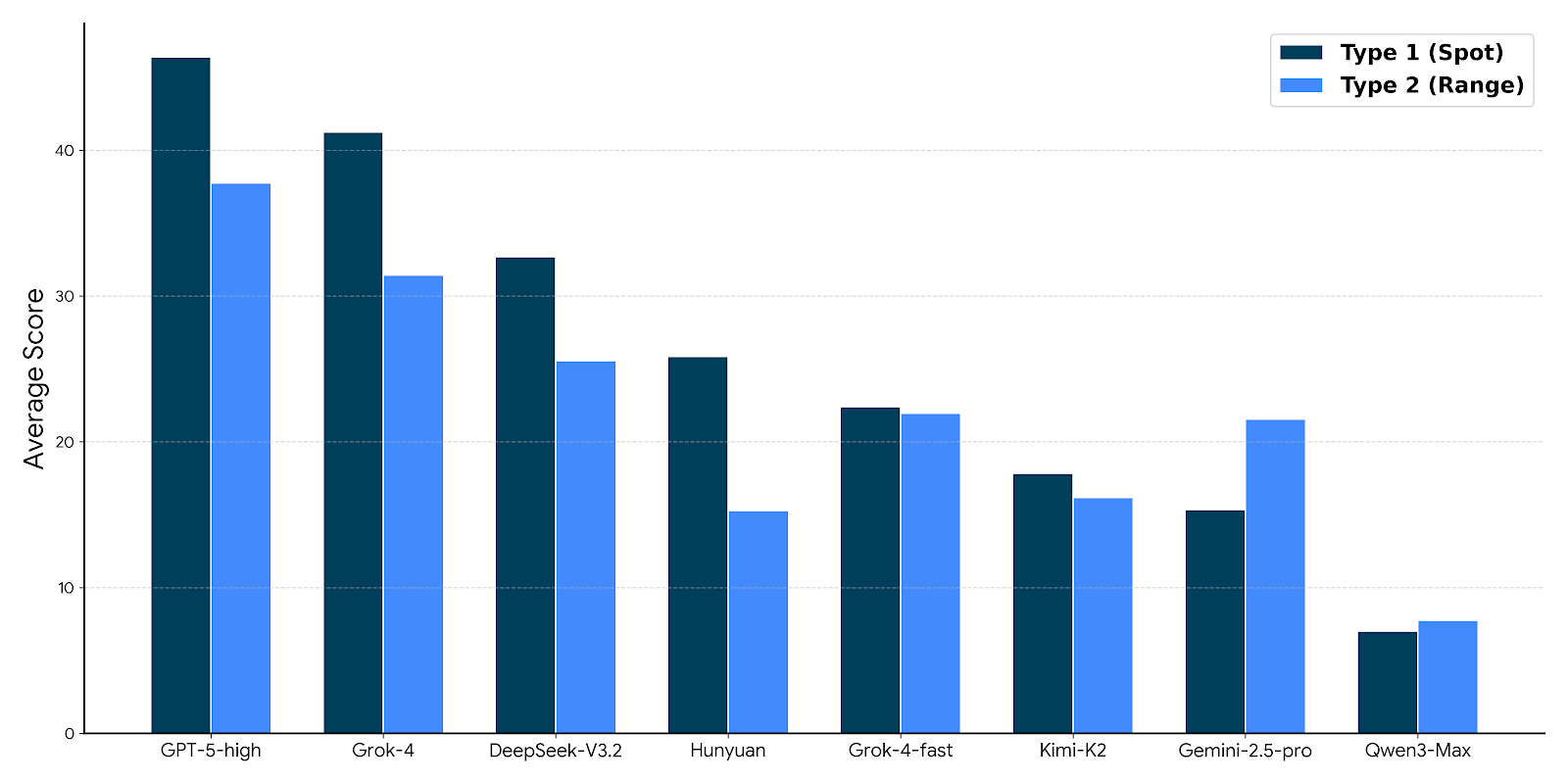}
    \caption{Results on FutureX-Finance between Oct.~24$^\text{th}$ and Nov.~28$^\text{th}$.}
    \label{fig:finance_results}
\end{figure}

\begin{itemize}
    \item Insight 1: Performance Tiers.
    The evaluation reveals distinct performance groups. GPT-5-high and Grok-4 lead the benchmark, achieving average Type 1 scores of 46.37 and 41.25, respectively. This suggests these models are more effective at narrowing down numerical predictions within the 5\% error tolerance. In comparison, models like Qwen3-Max and Kimi-K2 score significantly lower, often failing to meet the precision threshold required by the scoring function.

    \item Insight 2: Impact of Task Complexity.
    Performance generally declines when moving from Type 1 (Spot Prediction) to Type 2 (Window Extremum Prediction). For instance, Hunyuan sees its score drop from 25.85 in Type 1 to 15.29 in Type 2. This trend indicates that identifying extrema across a multi-day window is more challenging than forecasting a single point. However, DeepSeek-V3.2 maintains relatively stable performance across both task types, implying better consistency in handling temporal data.

    \item Insight 3: Metric Sensitivity and Model Limits.
    No model achieved an average score above 50. This low absolute performance highlights the difficulty of the benchmark. While models can process financial news, converting that information into a numerical forecast with less than 5\% error remains a significant challenge. The strict penalty function effectively filters out broad or vague estimates, rewarding only high-precision outputs.
\end{itemize}

\section{FutureX-Retail}
\label{sec:retail}

In the fast-paced e-commerce sector~\citep{min2025ecombench}, accurate demand forecasting is the engine of supply chain efficiency. \textbf{FutureX-Retail} simulates the role of an inventory analyst, requiring agents to \emph{predict product sales volumes} based on historical trends and market signals that they can gather. Unlike previous domains where information is publicly retrievable, this domain introduces a \emph{Data Scarcity Challenge}: historical sales data is often proprietary or ephemeral. Therefore, agents must synthesize provided internal snapshots (HTML) with external market intelligence to model consumer demand.

\subsection{Data Construction and Taxonomy}

The dataset focuses on the ultra-competitive cross-border e-commerce market, sourced from \texttt{Temu}, a rapidly growing global platform.

\textbf{Product Taxonomy.} \quad
The dataset covers \emph{240 distinct products} categorized into a hierarchical taxonomy of \emph{4 Major Categories} and \emph{24 Sub-categories} to ensure diversity in consumer behavior patterns.
\begin{itemize}
    \item \emph{Electronics:} High-volatility items including Headphones, Keyboards, and Smartwatches.
    \item \emph{Apparel and Accessories:} Seasonal items such as Women's Dresses, Men's T-Shirts, and Jewelry.
    \item \emph{Home and Living:} Durable goods like Furniture, Kitchen Utensils, and Lighting.
    \item \emph{Beauty and Personal Care:} Fast-moving consumer goods (FMCG) including Skincare, Makeup, and Baby Care.
\end{itemize}

\subsection{Task Configuration and Variations}
To diagnose agent capabilities across different levels of information density and uncertainty modeling, we design a $2 \times 3$ experimental matrix, resulting in six distinct task configurations (e.g., Task 1a, 2b).
Example tasks are shown in~\Cref{tab:retail}.

\textbf{Input Conditions (Information Density).}\quad
We vary the historical context provided in the prompt to test both static extraction and dynamic trend extrapolation capabilities.

\begin{itemize}
    \item \textbf{Type A: Snapshot-Only Context (Static).} \\
    The agent is provided \textit{only} with the HTML source code of the product page from 7 days prior ($T_{-7}$).
    \begin{itemize}
        \item \textit{Challenge:} The agent must deduce sales velocity solely from metadata embedded in the snapshot (e.g., "sold recently" tags, review growth) without a historical numerical baseline.
        \item \textit{Input:} HTML Snapshot at $T_{-7}$.
    \end{itemize}

    \item \textbf{Type B: Sparse Time-Series Context (Dynamic).} \\
    In addition to the HTML snapshot from $T_{-7}$, the agent is explicitly provided with the exact sales volume from 14 days prior ($T_{-14}$).
    \begin{itemize}
        \item \textit{Challenge:} This enables—and requires—the agent to perform trend extrapolation by calculating the sales delta between $T_{-14}$ and the estimated $T_{-7}$ state to forecast the target $T_{0}$.
        \item \textit{Input:} Sales Volume at $T_{-14}$ + HTML Snapshot at $T_{-7}$.
    \end{itemize}
\end{itemize}

\textbf{Output Definitions (Prediction Granularity).}\quad
We require agents to output predictions in three increasingly complex formats to assess their ability to quantify uncertainty.

\begin{itemize}
    \item Level 1: Deterministic Point Forecast (Task 1) \hfill \\
    The agent must output a single, definite integer representing the most likely sales volume.
    \begin{quote}
        \small\texttt{\{"predicted\_sales\_volume": 551\}}
    \end{quote}

    \item Level 2: Top-K Probabilistic Forecast (Task 2) \hfill \\
    The agent must identify the \emph{top three} most likely outcomes and assign a probability to each, summing to 1.0. This tests calibration on high-confidence candidates.
    \begin{quote}
        \small\texttt{\{"551": 0.50, "550": 0.48, "552": 0.02\}}
    \end{quote}

    \item Level 3: Full Distribution Forecast (Task 3) \hfill \\
    The agent must list \emph{all} plausible outcomes with their associated probabilities. This rigorously evaluates the agent's understanding of ``tail risk'' and the full scope of demand variance.
    \begin{quote}
        \small\texttt{\{"551": 0.50, "550": 0.26, "552": 0.13, ..., "549": 0.02\}}
    \end{quote}
\end{itemize}

\begin{table}[h!]
    \centering
    \caption{Example of 2$\times$ 3 tasks in FutureX-Retail.}
    \label{tab:retail}
    \scriptsize 
    \renewcommand{\arraystretch}{1.5} 
    \begin{tabularx}{\textwidth}{@{} l X X X X X X @{}}
    \toprule
    \textbf{Task Type} & \textbf{1a} & \textbf{1b} & \textbf{2a} & \textbf{2b} & \textbf{3a} & \textbf{3b} \\ 
    \midrule
    
    Example Core Task & \multicolumn{6}{>{\hsize=\dimexpr6\hsize+10\tabcolsep}X}{You are an intelligent agent that can predict future events. The event to be predicted is: "Predict the sales volume of the product \{***\} from the merchant \{***\} on Temu in the US region at 2025-12-05, 14:00 GMT+8."} \\ 
    \midrule
    
    Historical Context & 
    & 
    The sales volume at 2025-11-21, 14:00 GMT+8 was \{***\}. & 
    & 
    The sales volume at 2025-11-21, 14:00 GMT+8 was \{***\}. & 
    & 
    The sales volume at 2025-11-21, 14:00 GMT+8 was \{***\}. \\ 
    \midrule
    
    \multirow{2}{*}{Requirement} & 
    & 
    & 
    \multicolumn{2}{>{\hsize=\dimexpr2\hsize+2\tabcolsep}X}{List exactly THREE sales volumes you think are most likely. Give your prediction in the form of a probability distribution. List these three sales volumes in order of probability (from highest to lowest), along with their probabilities (two-decimal-place numbers). The three probabilities must sum to 1. I will evaluate your prediction based on whether the true result falls within your predicted sales volumes and the probability distribution you provided.} & 
    
    \multicolumn{2}{>{\hsize=\dimexpr2\hsize+2\tabcolsep}X}{Give your prediction in the form of a probability distribution. List ALL sales volumes you think are likely. Give these sales volumes in order of probability (from highest to lowest), along with their probabilities (two-decimal-place numbers). The probabilities must sum to 1. I will evaluate your prediction based on whether the true result falls within your predicted sales volumes and the probability distribution you provided.} \\ 
    
    \cmidrule{2-7} 
     & \multicolumn{6}{>{\hsize=\dimexpr6\hsize+10\tabcolsep}X}{\# Note: \newline 1) You need to find the sales volume of the product from the HTML information provided.\newline 2) You can use search tools to collect necessary information.\newline \# Output Format (STRICT)\newline Return **exactly two lines**, no extra text:} \\ 
     \midrule
    
    Output Format & \multicolumn{2}{>{\hsize=\dimexpr2\hsize+2\tabcolsep}X}{1) **Prediction line**: A JSON object enclosed within \textless{}prediction\textgreater{} tags, containing your final predicted sales volume.} & \multicolumn{4}{>{\hsize=\dimexpr4\hsize+6\tabcolsep}X}{1) **Prediction line**: A JSON object enclosed within \textless{}prediction\textgreater{} tags, containing your final predicted sales volumes and probabilities.} \\ 
    \midrule
    
    Output Example & \multicolumn{2}{>{\hsize=\dimexpr2\hsize+2\tabcolsep}X}{\textless{}prediction\textgreater \{\{"predicted\_sales\_volume": 551\}\} \textless{}/prediction\textgreater\newline \textless{}myreasoning\textgreater {[}{[}The ...{]}{]} \textless{}/myreasoning\textgreater{}} & \multicolumn{2}{>{\hsize=\dimexpr2\hsize+2\tabcolsep}X}{\textless{}prediction\textgreater \{\{"551": 0.50, "550": 0.48, "552": 0.02\}\} \textless{}/prediction\textgreater\newline \textless{}myreasoning\textgreater {[}{[}The ...{]}{]} \textless{}/myreasoning\textgreater{}} & \multicolumn{2}{>{\hsize=\dimexpr2\hsize+2\tabcolsep}X}{\textless{}prediction\textgreater \{\{"551": 0.50, "550": 0.26, "552": 0.13, "553": 0.09, "549": 0.02\}\} \textless{}/prediction\textgreater\newline \textless{}myreasoning\textgreater {[}{[}The ...{]}{]} \textless{}/myreasoning\textgreater{}} \\ 
    \bottomrule
    \end{tabularx}
\end{table}

Given the importance of historical data in the Retail domain, we mainly evaluate models under Task 1b, 2b, and 3b in the following.

\subsection{Evaluation Metrics}

Given the variable granularity of e-commerce data (e.g., exact sales counts vs. fuzzy ``10k+'' badges), we implement a hybrid scoring mechanism.
Let $y$ denote the Ground Truth (GT). 
\begin{itemize}
    \item If $y$ is \texttt{None}, the sample is excluded from evaluation.
    \item If the agent's response is empty or unparseable, the score is $0$.
\end{itemize}
For valid responses, the scoring is defined as follows:

\textbf{Base Scoring Function.}\quad
We define a scoring function $S(v, y)$ for a single predicted value $v$ against the ground truth $y$, conditional on the granularity of $y$.

\paragraph{Case 1: Exact Ground Truth ($y \le 999$).}
When high-precision data is available, we use a \emph{Linear Decay Metric} with strict tolerance thresholds $\epsilon = 0.05$. The score decays linearly based on the relative error:
\begin{equation}
    S(v, y) = \max \left( 0, 1 - \frac{|v - y|}{y \times \epsilon} \right)
\end{equation}
If the prediction deviates beyond the threshold $\epsilon$, the score drops to 0.

\paragraph{Case 2: Coarse Ground Truth ($y > 999$).}
When the platform only displays fuzzy badges (e.g., ``1.5k+''), we map $y$ to a closed interval range $R_y = [L, U]$ (e.g., ``1.5k+'' $\rightarrow [1500, 1599]$). The evaluation becomes a \emph{Binary Range Check}:
\begin{equation}
    S(v, y) = \mathbb{I}(v \in R_y) = 
    \begin{cases} 
    1 & \text{if } L \le v \le U \\
    0 & \text{otherwise}
    \end{cases}
\end{equation}

\textbf{Task-Specific Aggregation.}\quad
Based on the base function $S(v, y)$, we calculate the final score for each task type:

\paragraph{Task 1 (Deterministic Prediction).}
The agent outputs a single value $\hat{y}$. The score is simply defined as:
\begin{equation}
    \text{Score}_{\text{task1}} = S(\hat{y}, y)
\end{equation}

\paragraph{Task 2 and 3 (Probabilistic Prediction).}
The agent outputs a distribution of $N$ predicted values $\{v_1, ..., v_N\}$ with associated probabilities $\{p_1, ..., p_N\}$, where $\sum p_i = 1$. The final score is the \emph{Expected Score} under the agent's predicted distribution:
\begin{equation}
    \text{Score}_{\text{prob}} = \sum_{i=1}^{N} p_i \times S(v_i, y)
\end{equation}
This metric rewards agents that assign a  higher confidence probability to accurate values.

\subsection{Results}
We use our internal agent evaluation framework (with search tools) to comprehensively evaluate 18 leading LLMs, including GPT-5.1-High, GPT-5-High, GPT-5-Medium, Claude-Opus-4.1, Claude-Sonnet-4.5, Kimi-K2-0905, Kimi-K2-1104, Kimi-K2-thinking, Grok-4, DeepSeek-V3.1, Gemini-2.5-flash\footnote{Due to high API error rate, we did not include the results of Gemini-2.5-pro here.}, o3-High, GLM-4.6, GLM-4.5, DeepSeek-R1, and Qwen3-Max.

\begin{figure}[htbp]
    \centering
    \includegraphics[width=\linewidth]{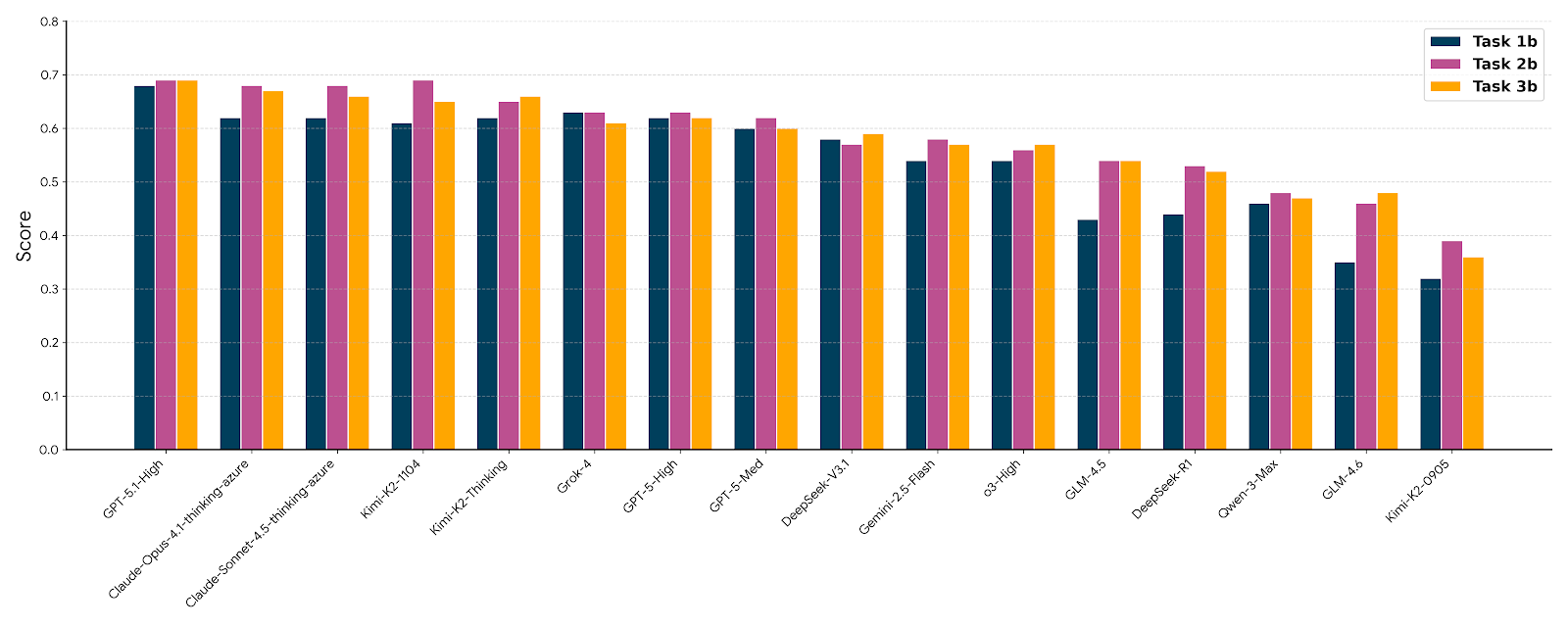}
    \caption{Results on FutureX-Retail between Nov.~12$^\text{th}$ and Dec.~3$^\text{rd}$.}
    \label{fig:retail_results}
\end{figure}

Figure \ref{fig:retail_results} shows the performance of leading LLMs on the FutureX-Retail, highlighting three critical trends as follows:

\begin{itemize}
    \item Insight 1: The ``Probabilistic Advantage''.
    Most models, including Claude-Opus and Kimi-K2, achieve higher scores in Task 2b (Top-3) and Task 3b (Full Distribution) compared to Task 1b (Point Prediction). This indicates that while agents may struggle to pinpoint an exact integer sales figure, they possess strong \emph{uncertainty calibration}. They successfully identify the ``high-probability neighborhood'' of the true value, which is often more valuable for inventory risk management than a brittle point estimate.
    \item Insight 2: Stability at the SOTA Frontier.
    GPT-5.1-High and Grok-4 exhibit remarkable stability across all three task types (Score $\approx 0.69$ and $0.63$ across 1b, 2b, and 3b). Unlike lower-tier models that show high variance between tasks, the SOTA model demonstrates a robust internal representation of the market, effectively collapsing the uncertainty distribution into a high-confidence prediction.
\end{itemize}

\subsubsection{Effects of Historical Context}
\begin{figure}[htbp]
    \centering
    \includegraphics[width=\linewidth]{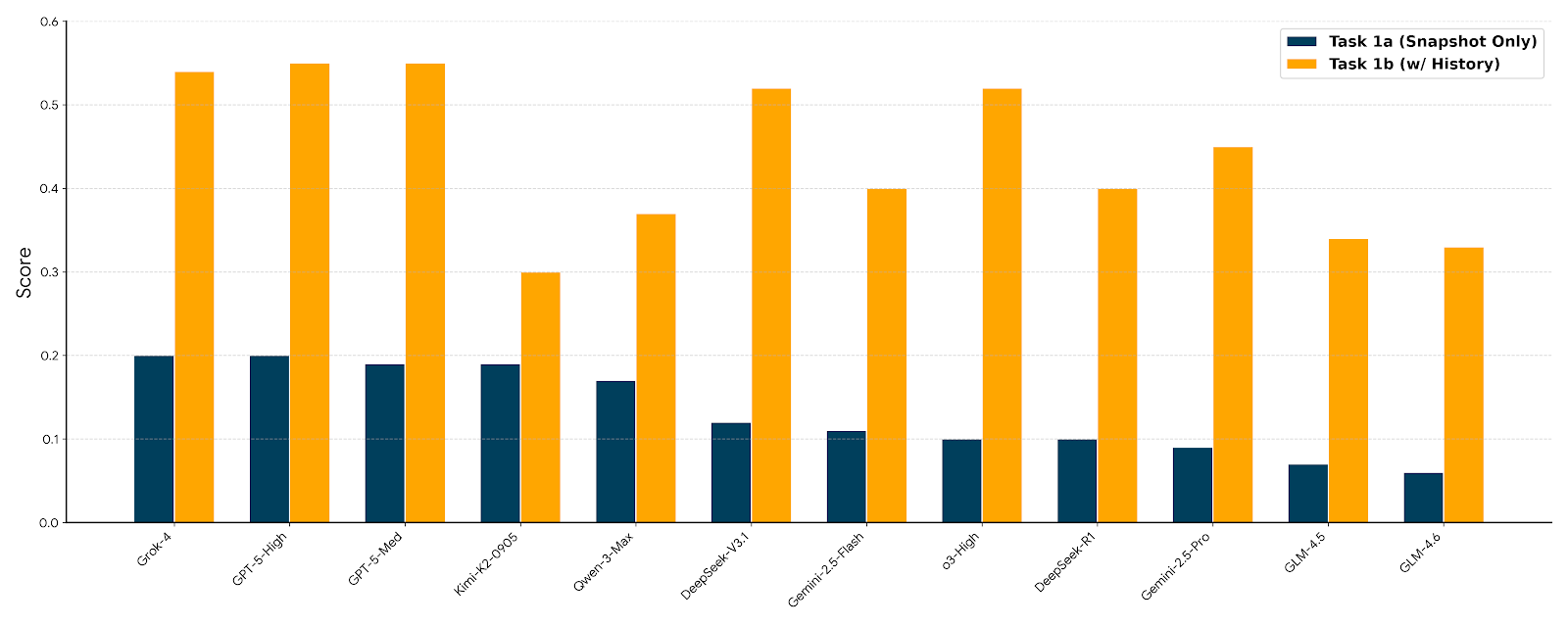}
    \caption{Comparison between Task 1a and Task 1b on FutureX-Retail.}
    \label{fig:retail_compare}
\end{figure}

To quantify the impact of historical context, we conduct a comparative analysis of model performance across Task 1a (snapshot only) and Task 1b (history-augmented). As illustrated in~\Cref{fig:retail_compare}, the inclusion of historical data yields substantial performance gains across all models. This empirical evidence validates the design of Type B tasks as a necessary benchmark for evaluating trend extrapolation capabilities.

\subsubsection{Analysis of In-Context Learning Sensitivity}

We further investigate how the characteristics of few-shot examples (demonstrations) influence model behavior. Our experiments reveal a significant \emph{positive correlation} between the provided examples and the model's output distribution across two key dimensions: \textit{Prediction Count} and \textit{Maximum Probability}.

\textbf{1. Sensitivity to Output Granularity (Prediction Count).}\quad
Models exhibit strong alignment with the granularity of the provided examples.
\begin{itemize}
    \item Single-Source Influence: Increasing the number of predicted values in the example directly drives the model to output a larger support set.
    \begin{itemize}
        \item When the example contains \emph{5 values}, the average output count is \emph{5.1} (Mode: 4.8).
        \item When the example is increased to \emph{8 values}, the average output count jumps to \emph{7.3} (Mode: 6.9), with the maximum count shifting from 9.6 to 14.0.
    \end{itemize}
    \item Mixed-Source Interpolation: When provided with multiple examples of varying granularity (e.g., one with 5 and one with 8 values), the model's behavior is \emph{interpolated but slightly biased towards the larger value}, producing a distribution centered between the two anchors.
\end{itemize}

\textbf{2. Sensitivity to Confidence Calibration (Max Probability).}\quad
The confidence level shown in the examples steers the model's own certainty, though with diminishing returns on average statistics.
\begin{itemize}
    \item Task 2 (Top-3 Forecast): As the example's max probability increases from $0.5 \rightarrow 0.7 \rightarrow 0.9$:
    \begin{itemize}
        \item The \textit{average} max probability of the model output increases marginally ($0.51 \rightarrow 0.56 \rightarrow 0.58$).
        \item However, the \textit{maximum} observed probability shifts significantly ($0.76 \rightarrow 0.76 \rightarrow 0.84$), indicating that strong examples unlock the model's potential for high-confidence predictions.
    \end{itemize}
    \item Task 3 (Full Distribution): A similar trend is observed. Boosting the example probability from $0.5 \rightarrow 0.7$ raises the model's average max probability from $0.39 \rightarrow 0.43$.
\end{itemize}

\textbf{3. Model-Specific Behaviors.}\quad
Different architectures manifest distinct ``personality traits'' under these conditions:
\begin{itemize}
    \item Grok-4 (The ``Long-Tail'' Specialist): It acts as a structural outlier, consistently generating a significantly higher number of predicted values (Average and Max) compared to all other models, suggesting a preference for exhaustive distribution coverage.
    \item GPT Series (The ``Egalitarian'' Reasoners): The GPT models exhibit a more balanced calibration. Their probability distributions tend to be centered with lower peak values, reflecting a more ``egalitarian'' or conservative approach to uncertainty quantification, avoiding overconfidence in any single outcome.
\end{itemize}

\section{FutureX-PublicHealth}
\label{sec:public_health}

In the wake of global health challenges~\citep{garrett2017challenge, albreht2023challenges}, the capacity to track infectious diseases is a cornerstone of societal resilience. \textbf{FutureX-PublicHealth} evaluates an agent's utility as a real-time biosurveillance tool. Unlike general queries, tasks in this domain require agents to navigate structured epidemiological reports, interpret demographic stratifications, and track pathogen dynamics with high temporal resolution.

\subsection{Data Construction and Sources}

To simulate the workflow of a public health analyst, this domain is strictly grounded in high-frequency official bulletins.

\textbf{Source Authority.}\quad Data is sourced exclusively from national public health institutes, specifically the \textit{Centers for Disease Control and Prevention} (US CDC) and the \textit{Chinese Center for Disease Control and Prevention} (China CDC). These sources represent the authoritative standard for respiratory virus surveillance.

\begin{figure}[htbp]
    \centering
    \includegraphics[width=\linewidth]{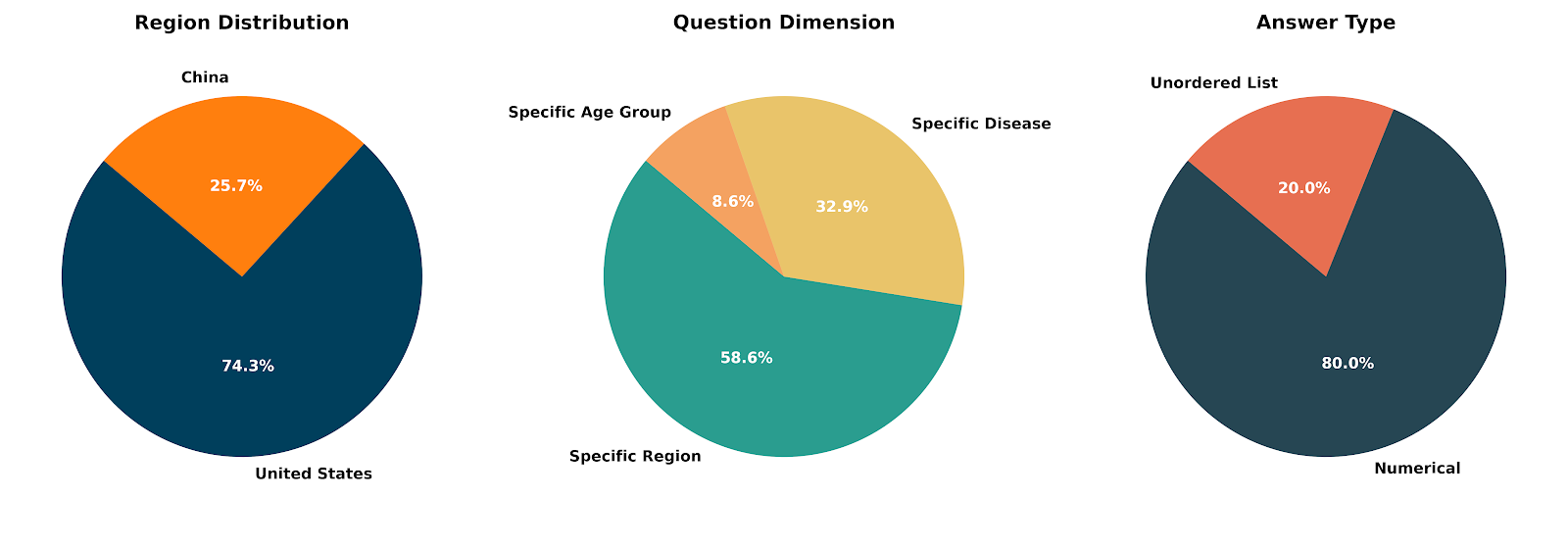}
    \caption{Statistics of FutureX-PublicHealth. The benchmark focuses on high-frequency weekly updates across different epidemiological dimensions.}
    \label{fig:health_distribution}
\end{figure}

\textbf{Scale and Granularity.}\quad The dataset consists of \textit{70 recurring weekly event templates}, expanded into \textit{382 specific variables}. A defining feature of this domain is its high temporal frequency; all tasks are updated on a weekly basis, requiring agents to distinguish between the latest reporting period and historical data. As detailed in~\Cref{fig:health_distribution}, the tasks cover diverse dimensions including specific diseases, geographic regions, and demographic groups (e.g., age 0-4).

\textbf{Task Formulation.}\quad Tasks in FutureX-PublicHealth require parsing hierarchical data structures, such as viral lineage breakdowns or age-stratified pathogen positivity rates. Here we showcase two task templates:

\paragraph{Example 1: Viral Subtyping (US CDC).}
\begin{quote}
    \textit{``\{Date\} according to the data from public health laboratories in the influenza surveillance report (FluView) of the US CDC, what will be the number of positive cases of \{t\}?''}
\end{quote}
\begin{itemize}
    \item The variable $t$ is chosen from \{A (H3), A (H1N1) pdm09, A (Subtyping not Performed), B (Victoria Lineage), B (Lineage Unspecified)\}.
    \item This task tests the agent's ability to distinguish between specific viral subtypes (e.g., H3 vs. H1N1 pdm09) within a dense laboratory report.
\end{itemize}

\paragraph{Example 2: Demographic Analysis (China CDC).}
\begin{quote}
    \textit{``\{Date\} according to China CDC data, among the 0-4 age group, what will be the top three pathogens with the highest positive rates of nucleic acid detection in \{t\} respiratory tract samples?''}
\end{quote}
\begin{itemize}
    \item The variable $t$ is chosen from \{Outpatient influenza-like cases, hospitalized severe acute respiratory infection cases\}.
    \item This task requires cross-referencing pathogen prevalence with demographic data to extract an ordered or unordered list of risks.
\end{itemize}

\subsection{Evaluation Metrics}
Given the precision required, we implement strict scoring protocols tailored to the answer type.

\textbf{Numerical Prediction Scoring.}\quad We employ a \emph{Relative Error Scoring} mechanism. Let $y$ be the ground truth value and $\hat{y}$ be the agent's prediction. We define a strict tolerance threshold $\epsilon \in \{0.05, 0.10\}$ depending on the variable's volatility. The score $S_{\text{num}}$ is calculated as:

\begin{equation}
    S_{\text{num}} = 
    \begin{cases} 
    0 & \text{if } \frac{|\hat{y} - y|}{y} > \epsilon \\
    1 - \frac{|\hat{y} - y|}{y \times \epsilon} & \text{if } \frac{|\hat{y} - y|}{y} \leq \epsilon
    \end{cases}
\end{equation}

Any prediction deviating beyond the threshold $\epsilon$ receives a score of 0, strictly penalizing hallucinated precision.

\textbf{Unordered List/Classification Scoring.}\quad For tasks involving categorical states or sets (e.g., list of affected provinces), we use a \emph{Set-Based Overlap Metric}. Let $G$ be the set of ground truth elements and $P$ be the set of predicted elements:

\begin{equation}
    S_{\text{cat}} = 
    \begin{cases} 
    1.0 & \text{if } P = G \text{ (Exact Match)} \\
    0.5 & \text{if } P \subset G \text{ or } P \cap G \neq \emptyset \text{ (Partial Match)} \\
    0.0 & \text{otherwise (Failure)}
    \end{cases}
\end{equation}

This rigorous evaluation ensures that high scores reflect true domain grounding rather than probabilistic guessing.

\subsection{Results}
\begin{figure}[htbp]
    \centering
    \includegraphics[width=\linewidth]{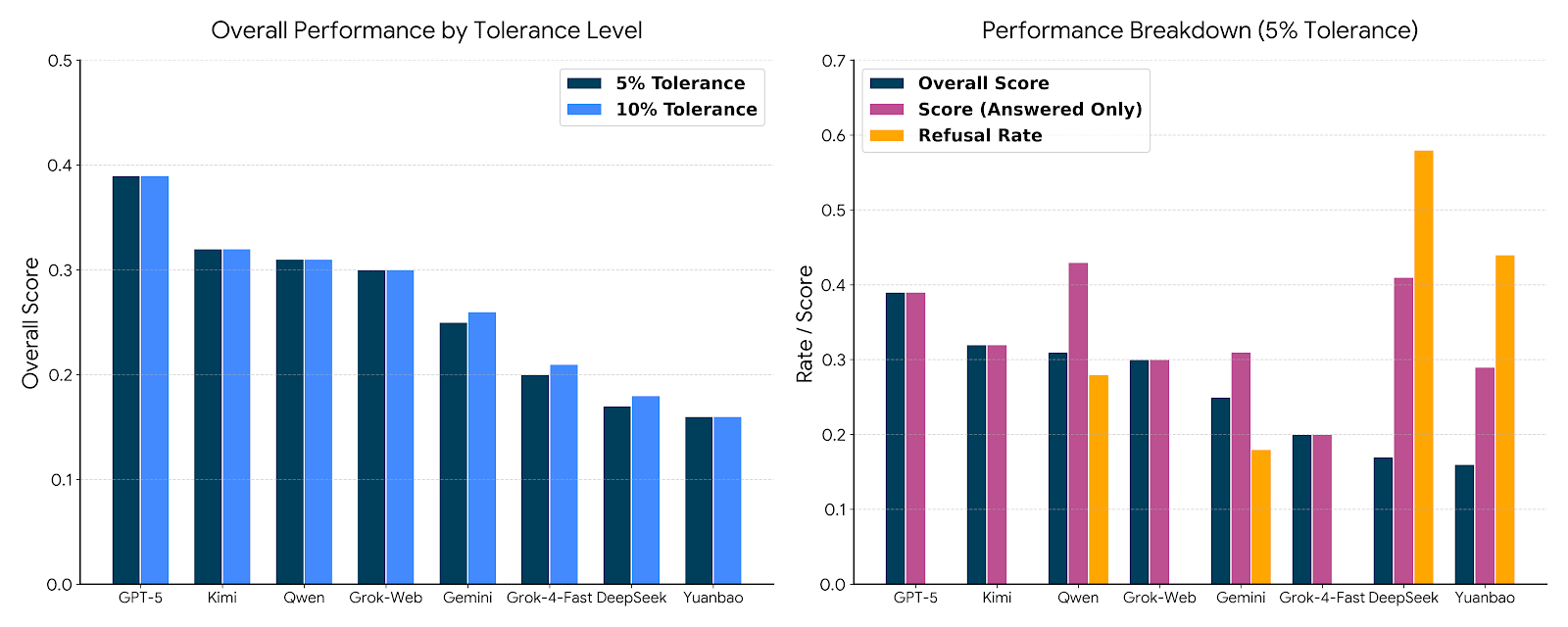}
    \caption{Results on FutureX-PublicHealth between Nov.~14$^\text{th}$ and Nov.~28$^\text{th}$.}
    \label{fig:health_results}
\end{figure}

We consider 8 leading agentic LLMs: GPT-5-High, Grok-4, Kimi-K2-thinking, Gemini-2.5-pro, and Grok-4-fast (evaluated with tool-use), alongside Qwen3-Max, DeepSeek-V3.2-Exp, and Yuanbao (evaluated with web-browsing). 
Performances on data between Nov. 14$^\text{th}$ and Nov. 28$^\text{th}$ are summarized in~\Cref{fig:health_results}.

Based on the performance metrics in the Public Health domain, we identify three critical observations regarding agent reliability and the precision-recall trade-off.

\begin{itemize}
    \item \textbf{Insight 1: The Precision-Recall Trade-off.} 
    A striking reversal occurs when evaluating only answered queries. While GPT-5-High and Kimi-K2-thinking lead the overall leaderboard due to their high coverage (0\% refusal), Qwen3-Max and DeepSeek-V3.2-Exp significantly outperform them in pure accuracy on answered tasks.
    \begin{itemize}
        \item If refusals are excluded from the denominator, Qwen3-Max and DeepSeek-V3.2-Exp ascend to the $1^{st}$ and $2^{nd}$ positions, surpassing GPT-5-High. 
        \item This indicates that Qwen3-Max and DeepSeek-V3.2-Exp possess superior \emph{domain precision} for high-confidence samples but suffer from an overly conservative retrieval threshold. They function as ``High-Precision Specialists'' rather than ``Generalist Monitors.''
    \end{itemize}

    \item \textbf{Insight 2: Divergent Safety Strategies in Medical Contexts.} 
    The refusal rates in Public Health are notably higher than in Natural Disasters for certain models, reflecting stricter safety alignment for medical/health topics. 
    \begin{itemize}
        \item DeepSeek-V3.2-Exp exhibits an extreme refusal rate of nearly 60\%, followed by Yuanbao (44\%). 
        \item In contrast, GPT-5-High, Kimi-K2-thinking, and Grok-4 maintain a 0\% refusal rate. 
        \item This bifurcation highlights a strategic difference: Aggressive agents prioritize utility (providing a best-effort estimate), while conservative agents prioritize safety (avoiding potential misinformation in health contexts at the cost of utility).
    \end{itemize}
\end{itemize}

\section{FutureX-NaturalDisaster}
\label{sec:natural_disaster}

As climate volatility intensifies, the capability to anticipate environmental hazards becomes a critical safety imperative~\citep{yu2018big, ritchie2022natural}. \textbf{FutureX-NaturalDisaster} assesses an agent's ability to function as a biosurveillance and environmental monitoring tool. This domain requires agents to synthesize multi-modal signals (e.g., meteorological data, geological reports) and perform rigorous spatial-temporal reasoning to forecast hazard trajectories and impacts.

\subsection{Data Construction and Sources}

To ensure the reliability required for high-stakes safety scenarios, FutureX-NaturalDisaster is grounded exclusively in authoritative data from national and international government agencies.

\textbf{Source Authority.}\quad We aggregate real-time data from official bodies, including the \textit{National Oceanic and Atmospheric Administration} (NOAA)\footnote{\url{https://www.drought.gov/}}, the \textit{U.S. Geological Survey} (USGS)\footnote{\url{https://www.usgs.gov/programs/VHP/}}, the \textit{National Meteorological Center of China} (NMC)\footnote{\url{https://www.nmc.cn/}}, and the \textit{UK Met Office}.

\begin{figure}[htbp]
    \centering
    \includegraphics[width=\linewidth]{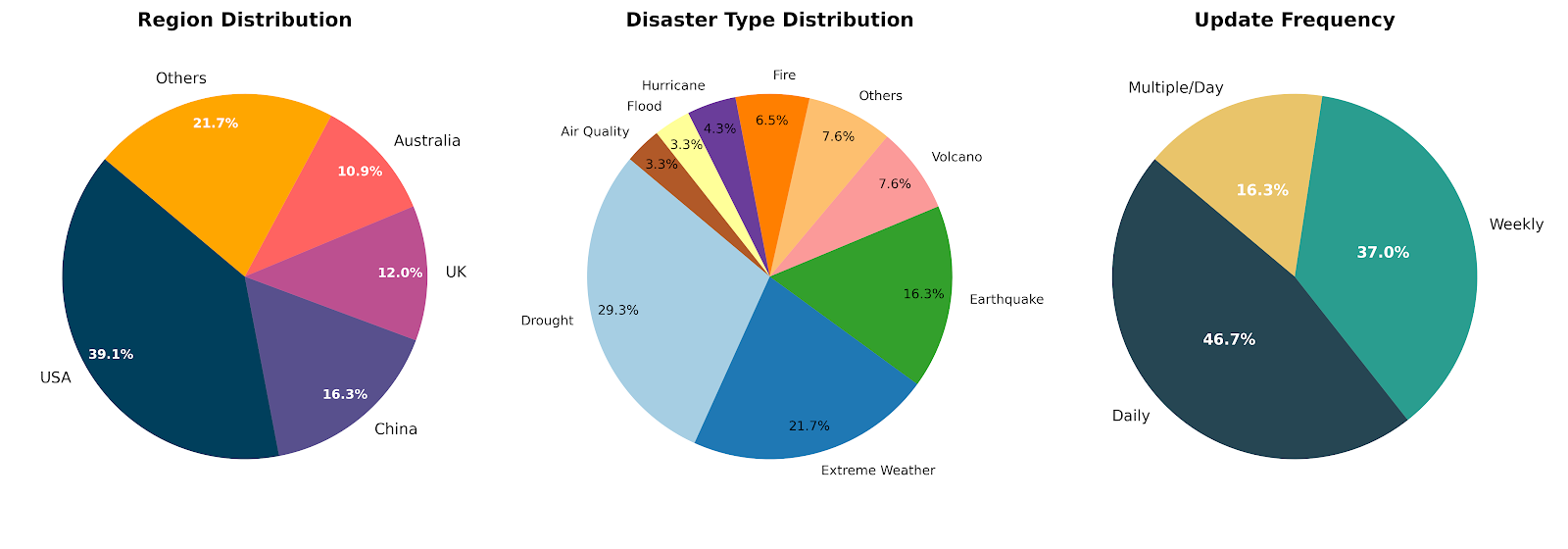}
    \caption{Statistics of FutureX-NaturalDisaster. The dataset covers diverse regions and disaster types with varying update frequencies.}
    \label{fig:disaster_stats}
\end{figure}

\textbf{Scale and Diversity.}\quad The dataset comprises \emph{92 distinct event templates} instantiated with \emph{446 specific variables}. As shown in Figure~\ref{fig:disaster_stats}, the benchmark achieves global coverage across 15 regions and covers a wide spectrum of hazard types, ranging from high-frequency events like earthquakes to slow-onset phenomena such as droughts.

\textbf{Task Formulation.}\quad Tasks in this domain simulate the workflow of a risk analyst monitoring live data streams. They are generally categorized into \textit{Numerical Prediction} and \textit{State Classification}.
Here we showcase two task templates:

\paragraph{Example Template 1: Numerical Prediction (Drought Monitoring).}
\begin{quote}
    \textit{``\{Date\} according to NOAA's National Integrated Drought Information System, what percentage of the U.S. will be classified as \{t\}}?''
\end{quote}
\begin{itemize}
    \item The variable $t$ is chosen from \{Abnormally Dry, Moderate Drought, Severe Drought, Extreme Drought, Exceptional Drought\}.
    \item The agent must interpret quantitative reports and output a precise percentage value.
\end{itemize}

\paragraph{Example 2: State Classification (Cyclone Warning).}
\begin{quote}
    \textit{``\{Date\} according to the Tropical Cyclone Outlook issued by TCWC Jakarta, the potential growth of tropical cyclones in the next \{$t_1$\} hours is predicted as \{$t_2$\}?''}
\end{quote}
\begin{itemize}
    \item The variable $t_1$ is chosen from \{24, 48, 72\}.
    \item The variable $t_2$ is chosen from \{Low/Medium/High\}.
    \item The agent must locate the specific regional bulletin and extract the probabilistic forecast.
\end{itemize}

\subsection{Evaluation Metrics}

We apply the same rigorous scoring protocols as defined in the Public Health domain (\Cref{sec:public_health}) to ensure consistency across high-stakes verticals.

\textbf{Numerical Prediction.}\quad For quantitative indicators (e.g., case counts), we use the \emph{Relative Error Scoring} function with strict tolerance thresholds ($\epsilon \in \{0.05, 0.10\}$). The score decays linearly as the error approaches the threshold and drops to zero immediately beyond it.

\textbf{List Generation.}\quad For tasks requiring the identification of top pathogens (Unordered List), we employ the \emph{Set-Based Overlap Metric}:
\begin{itemize}
    \item \emph{1.0 Point:} Exact match (all correct elements identified, no hallucinations).
    \item \emph{0.5 Points:} Partial match (subset of correct elements or inclusion of incorrect ones).
    \item \emph{0.0 Points:} Failure to retrieve valid data.
\end{itemize}

\subsection{Results}
\begin{figure}[htbp]
    \centering
    \includegraphics[width=\linewidth]{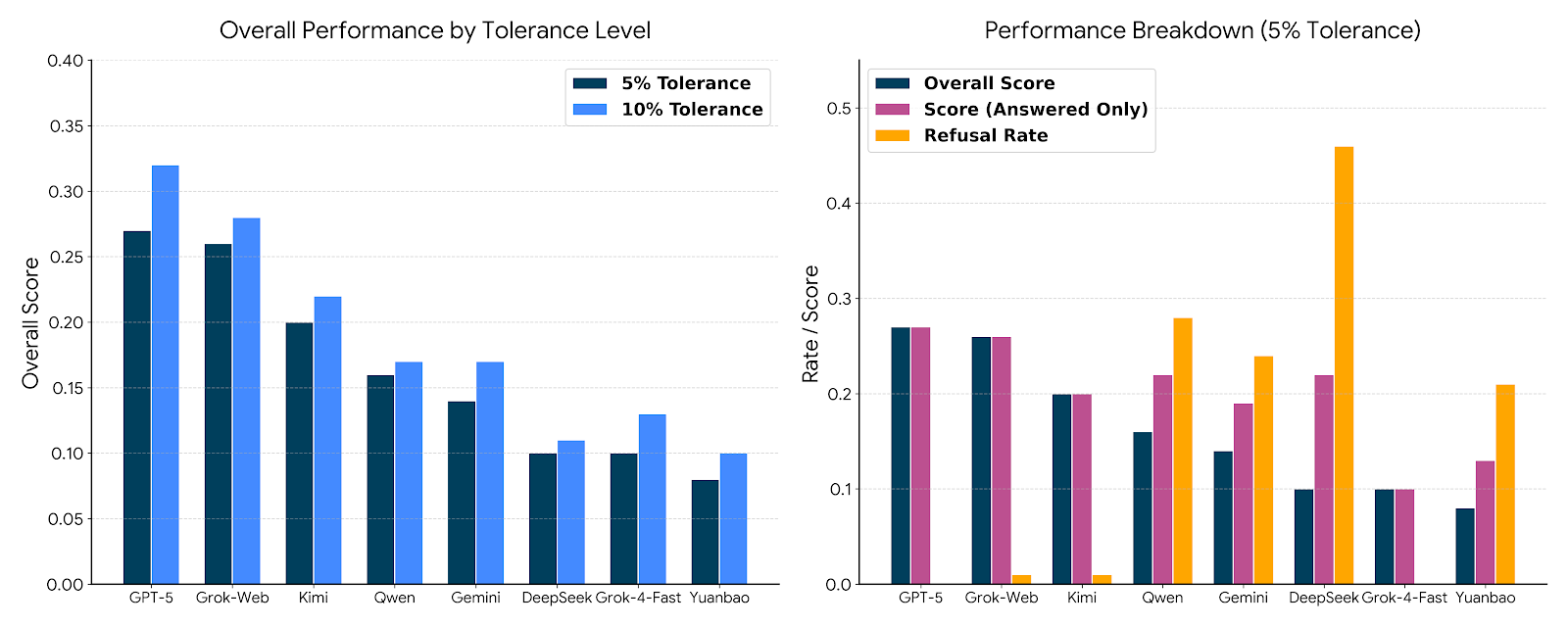}
    \caption{Results on FutureX-NaturalDisaster between Oct.~24$^\text{th}$ and Nov.~28$^\text{th}$.}
    \label{fig:disaster_result}
\end{figure}

We consider 8 leading agentic LLMs: GPT-5-High, Grok-4, Kimi-K2-thinking, Gemini-2.5-pro, and Grok-4-fast (evaluated with tool-use), alongside Qwen3-Max, DeepSeek-V3.2-Exp, and Yuanbao (evaluated with web-browsing). 
Performances on data between Oct. 24$^\text{th}$ and Nov. 28$^\text{th}$ are summarized in~\Cref{fig:disaster_result}.

Based on the quantitative results in the Natural Disaster domain, we derive four key insights regarding model behavior and reliability.
\begin{itemize}
    \item \textbf{Insight 1: Dominance of Aggressive Retrieval Agents.} 
    GPT-5, and Grok-web establish a clear lead with negligible refusal rates ($\le 1\%$). Crucially, since these events are grounded in objective government data (e.g., NOAA, USGS), the success of these models proves that the answer information \textit{is} accessible online. Their high performance demonstrates superior \emph{Information Retrieval (IR) Recall}, enabling them to locate specific data points that other models fail to find.
    \item \textbf{Insight 2: The Refusal Divergence.} 
    We observe a stark contrast in LLM behavior. While top-tier models attempt nearly every query, a cohort including DeepSeek (46\%), Qwen (28\%), and Gemini (24\%) exhibits significant refusal rates. This bifurcation suggests a fundamental difference in how agents handle the ``Unknown'': aggressive agents push to find a plausible answer, while others default to abstaining when high-confidence evidence is not immediately retrieved.

    \item \textbf{Insight 3: Retrieval-Induced Refusal vs. Safety Alignment.} 
    While refusal is often attributed to safety alignment, our error analysis suggests that \emph{Search Insufficiency} is a primary driver. 
    \begin{itemize}
        \item Since the ground truth data exists publicly (proven by GPT-5's success), a claim of ``Data not available'' (common in Qwen/Yuanbao) indicates a \emph{False Negative in Retrieval}. 
        \item Therefore, the lower ranking of these models stems partly from an inability to navigate deep-web government databases effectively, rather than solely from a conservative safety policy. When the search component fails to fetch the precise report, the reasoning component is forced to refuse.
    \end{itemize}

    \item \textbf{Insight 4: Taxonomy of Refusal Patterns.} 
    Qualitative analysis of refusal responses distinguishes between two failure modes:
    \begin{itemize}
        \item \emph{Capability-Based Refusal (Search Failure):} Models like Qwen and Yuanbao frequently cite data unavailability (e.g., \textit{``No authoritative data exists''}), which objectively reflects a failure in the search pipeline to locate existing reports.
        \item \emph{Policy-Based Refusal (Uncertainty Avoidance):} Models like DeepSeek and Gemini tend to use definitive negation (e.g., \textit{``Unable to predict''}), suggesting a mechanism that suppresses output when the retrieved context does not meet an internal confidence threshold.
    \end{itemize}
\end{itemize}
\section{FutureX-Search: From Prediction to Retrieval}
\label{sec:search}

While the core of FutureX focuses on forecasting, the reliability of any predictive agent is fundamentally grounded in its ability to accurately retrieve historical data. \textbf{FutureX-Search} serves as a control experiment, transforming ``past'' prediction tasks (where ground truth is now established) into rigorous information retrieval benchmarks. This domain evaluates whether agents can effectively navigate the search space when the answer already exists.

\subsection{Dataset Construction: Temporal Retrofitting}

We transform high-quality prediction questions from the FutureX~\cite{zeng2025futurex} archive into search queries. The construction pipeline consists of three phases: Calibration, Transformation, and Obfuscation.

\subsubsection{Phase 1: Discriminative Calibration}
To ensure the benchmark effectively differentiates model capabilities, we apply a strict filtering mechanism based on the performance of 23 baseline models on the original prediction tasks.
\begin{itemize}
    \item \textbf{Exclusion of Trivia:} Questions answered correctly by $\ge 19$ models are removed (Too Easy).
    \item \textbf{Exclusion of Noise:} Questions answered correctly by $\le 3$ models are removed (Too Hard/Ambiguous).
\end{itemize}
This results in a ``Goldilocks'' dataset where the information is retrievable but requires non-trivial search efforts.

\subsubsection{Phase 2: Temporal Shift (Standard Search)}
The filtered questions are syntactically transformed from future tense to past tense to form the FutureX-Search dataset ($N=100$).

\begin{quote}
    Original (FutureX): ``On 2025-09-04, \textit{what will be} the central parity rate...?'' \\
    \centering$\downarrow$ \\
    Standard Search (FutureX-Search): ``On 2025-09-04, \textit{what was} the central parity rate...?''
\end{quote}

\subsubsection{Phase 3: Complexity Injection (Complex Search)}
To evaluate \emph{Multi-hop Reasoning} and \emph{Entity Resolution}, we further process a subset of questions ($N=39$) by introducing ``Descriptive Entity Masks''. This process involves:
\begin{enumerate}
    \item Seed Extraction: Identifying a pivot entity in the query (e.g., \textit{``Japanese Yen''}).
    \item Wiki-Mask Generation: Generating a descriptive riddle based on Wikipedia knowledge (e.g., \textit{``There is a certain currency... The country where this currency is used...''}).
    \item Fusion: Combining the mask with the query to create FutureX-Search.
\end{enumerate}
This allows us to construct search queries (Level 3) similar to those in BrowseComp~\cite{wei2025browsecomp}, evaluating both the search and resaoning capabilities of LLMs.
As shown in~\Cref{fig:search}, this creates more complicated search tasks where all LLMs' performances drop a lot.
Details of the whole complexity injection process can be found in~\citet{qiu2025bmgq}.

\subsection{Task Formulation}

FutureX-Search presents three distinct levels of query complexity, requiring agents to bridge the gap between implicit descriptions and explicit data retrieval.

\paragraph{Level 1: Direct Retrieval.}
The agent receives a clear, explicit query about a historical event.
\begin{quote}
    \textit{Query:} ``On 2025-09-04, what was the central parity rate of the RMB against 100 Japanese Yen?''
\end{quote}

\paragraph{Level 2: Explicit Entity Resolution (Type 1).}
The agent must first identify the masked entity and then retrieve the data. The prompt explicitly asks ``What is this currency?''.
\begin{quote}
    \textit{Query:} ``[Description of Currency]... \emph{What is this currency?} On 2025-09-04, what was the rate of RMB against 100 of this currency?''
\end{quote}

\paragraph{Level 3: Implicit Multi-hop Retrieval (Type 2).}
The agent must internally resolve the masked entity to perform the search, without being explicitly prompted to identify it. This tests the agent's ability to handle ambiguous references.
\begin{quote}
    \textit{Query:} ``[Description of Currency]... On 2025-09-04, what was the rate of RMB against 100 of this currency?''
\end{quote}

\subsection{Evaluation Metrics}
Since these are fact-based retrieval tasks with established ground truth, we employ \emph{Exact Match (EM)} and \emph{F1 Score}\footnote{Since sometimes models refuse to answer a question.} to evaluate the accuracy of the retrieved numerical values and entities.
For individual questions, we use 0/1 loss.
For Type 1 questions (Level 2), a correct answer to the first question awards 0.4 points (0 for incorrect), while a correct answer to the second question awards 0.6 points (0 for incorrect).

\subsection{Results}
\begin{figure}
    \centering
    \includegraphics[width=\linewidth]{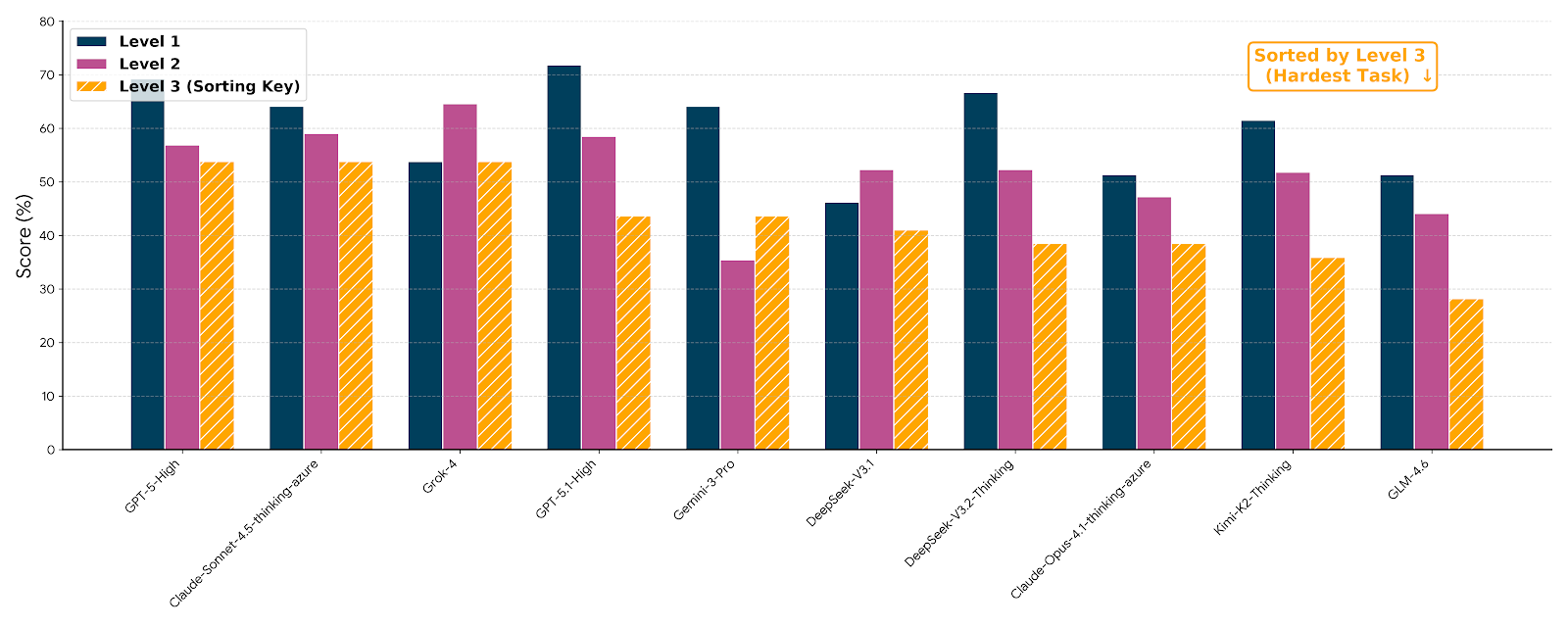}
    \caption{Results on FutureX-Search.}
    \label{fig:search}
\end{figure}

Based on our internal agent evaluation framework, we test different LLMs with search tools, including GPT-5.1-High, GPT-5-High, Claude-Opus-4.1, Claude-Sonnet-4.5, Grok-4, Gemini-3-pro, DeepSeek-V3.1, DeepSeek-V3.2, Kimi-K2-thinking, and GLM-4.6.
Results are shown in~\Cref{fig:search}.

We find that:
\begin{itemize}
    \item \textbf{Insight 1: Validation of Task Complexity.}
    The steep performance degradation observed in most models validates the difficulty of Level 3 (Implicit Multi-hop) tasks. 
    \begin{itemize}
        \item Models that excel at direct retrieval (Level 1) often falter when entity resolution is required. For instance, GPT-5.1-High achieves a dominant 71.8\% on Level 1 but drops significantly to 43.6\% on Level 3.
        \item This $\sim$28-point drop confirms that \textit{knowing} a fact (retrieval) is distinct from \textit{deducing} which fact to find (reasoning). The Level 3 task successfully filters out ``naive searchers'' from ``reasoning agents''.
    \end{itemize}

    \item \textbf{Insight 2: The ``Invariance'' of Grok-4.}
    A remarkable anomaly is observed in Grok-4. Unlike other models that follow a downward trend ($L1 > L2 > L3$), Grok-4 maintains perfect stability, scoring 53.8\% across Level 1 and Level 3.
    \begin{itemize}
        \item While its direct retrieval capability (Level 1) is mediocre compared to GPT-5.1, its ability to handle complex, implicit queries (Level 3) is top-tier.
        \item This suggests Grok-4's architecture prioritizes deep semantic alignment over raw knowledge indexing, making it uniquely robust against query obfuscation.
    \end{itemize}

    \item \textbf{Insight 3: The SOTA Ceiling.}
    Despite the capabilities of current flagship models, Level 3 performance hits a ``glass ceiling'' at approximately 54\% (shared by Grok-4, Claude-Sonnet-4.5, and GPT-5-High).
    \begin{itemize}
        \item No model approaches the 70\%+ accuracy seen in simple retrieval tasks. 
        \item This indicates that Implicit Entity Resolution—the ability to identify a search target from a vague description without explicit instruction—remains an unsolved challenge and a critical frontier for the next generation of Search Agents.
    \end{itemize}
\end{itemize}
\section{Related Work} \label{sec:related}

This section reviews the works on LLM agent evaluation, including general-purpose agent capabilities, dynamic future prediction, and domain-specific vertical evaluation. 
We highlight how FutureX-Pro builds upon the methodological foundation of previous work, FutureX~\cite{zeng2025futurex}, while addressing the critical need for deep, expert-level reasoning in high-stakes domains.

\paragraph{General-Purpose Agent Benchmarks.} 
The initial wave of agentic benchmarks focuses on evaluating broad, fundamental capabilities such as tool usage, planning, and web navigation in open-domain settings. 
Benchmarks like AgentBench~\citep{liu2024agentbench} and GAIA~\citep{mialon2023gaia} assess an agent's ability to act as a general-purpose assistant, solving tasks that are conceptually simple for humans but require complex multi-step reasoning for models. Similarly, WebArena~\citep{zhou2024webarena} and Mind2Web~\citep{deng2023mind2web} evaluate agents in high-fidelity simulated web environments, testing their ability to execute long-horizon interaction tasks. While these benchmarks effectively measure foundational agentic skills, they primarily operate on static problems with predefined scopes. They do not evaluate the ability to synthesize heterogeneous data streams for forward-looking analysis, a gap that necessitated the development of dynamic evaluation frameworks.

\paragraph{Dynamic Evaluation and Future Prediction.} 
To address the limitations of static evaluation—specifically data contamination and the lack of temporal reasoning—recent works have pivoted toward dynamic and live benchmarks. FutureX~\citep{zeng2025futurex} proposes the ``contamination-impossible by design'' architecture, establishing a pipeline that evaluates agents on their ability to predict verifiable future events. By leveraging a massive set of diverse, open-web questions (spanning sports, entertainment, and general politics), FutureX demonstrates that agents could serve as effective generalist forecasters. Other works, such as ForecastBench~\citep{kargerforecastbench} and LiveBench~\citep{white2024livebench}, also tackle the challenge of evaluating models on non-static data. However, these benchmarks, including the original FutureX, predominantly prioritize \emph{breadth} over \emph{depth}. They evaluate an agent's general adaptability to new information but often lack the specificity required to test expert-level reasoning. For instance, predicting a sports match outcome (a common task in FutureX) relies on aggregating public sentiment and stats, whereas forecasting an epidemic curve requires interpreting highly technical, domain-specific indicators—a capability that generalist benchmarks fail to isolate.

\paragraph{Vertical Domain Evaluation: From Breadth to Depth.} 
The transition from generalist assistants to specialist agents requires benchmarks rooted in vertical expertise. Existing domain-specific benchmarks, such as FinQA~\citep{chen2021finqa} and FinSearchComp~\citep{hu2025finsearchcomp} for finance or MedQA~\citep{jin2020disease} for medicine, focus on static knowledge retrieval and reasoning over historical documents. While effective for testing domain knowledge, they do not assess an agent's \emph{predictive utility} in real-world scenarios. FutureX-Pro bridges the gap between the dynamic evaluation pipeline of FutureX and the rigorous demands of vertical domains. Unlike FutureX, which aggregates widely accessible public information, FutureX-Pro focuses on \emph{high-value verticals}—Finance, Retail, Public Health, and Natural Disaster. These domains impose distinct challenges regarding source authority and data interpretation. For example, in the Finance domain of FutureX-Pro, agents must not only read news but specifically interpret corporate disclosures to forecast market movement, contrasting with the general news synthesis in FutureX. By shifting the focus from general breadth to high-stakes depth, FutureX-Pro evaluates whether agents can transcend generic web search to function as reliable junior analysts in specialized fields.

\section{Conclusion and Future Outlook}

This report introduces \textbf{FutureX-Pro}, a specialized framework designed to evaluate agentic capabilities within high-value vertical domains—Finance, Retail, Public Health, and Natural Disaster. Moving beyond generalist open-domain tasks or speculative forecasting benchmarks, FutureX-Pro explicitly shifts the evaluation focus to the rigorous demands of sectors that dictate economic and societal stability.

Our investigation highlights a fundamental distinction between ``conversational fluency'' and ``domain competence'':

\begin{itemize}
    \item \textbf{Source Authority and Grounding:} Unlike generalist benchmarks that may rely on crowd wisdom or broad news aggregation, FutureX-Pro mandates that agents ground their reasoning in authoritative, heterogeneous data sources—from financial disclosures to meteorological charts. Our findings in the Public Health and Natural Disaster domains reveal that performance bottlenecks often stem from ``Search Insufficiency''—the inability to effectively retrieve data from deep-web government databases—rather than simple reasoning failures.
    
    \item \textbf{Precision Beyond Plausibility:} In high-value domains, a plausible answer is insufficient if it lacks quantitative rigor. The strict evaluation in Finance (requiring a relative error of less than 5\%) exposes a significant gap: while agents can process qualitative news, they struggle to translate this into the precise numerical forecasts required for high-stakes environments.
    
    \item \textbf{The Value of Uncertainty:} Our analysis in the Retail domain challenges the notion that ``prediction'' equals ``certainty.'' The results demonstrate that agents offer greater utility through a ``Probabilistic Advantage''—identifying high-probability neighborhoods of outcomes—rather than being forced into brittle deterministic estimates.
\end{itemize}

To advance agentic systems from general-purpose assistants to capable partners in high-value domains, future research could address three critical frontiers:

\begin{itemize}
    \item \textbf{From Surface Retrieval to Deep Domain Grounding:} The reliability of an agent is defined by its information supply chain. Future work must move beyond generic web search to encompass ``Reasoning-First'' Retrieval strategies capable of navigating authoritative, structured data sources. As shown in FutureX-Search, the ability to perform implicit entity resolution and deduce \textit{what} to search for is a prerequisite for success in complex verticals.
    
    \item \textbf{Standardizing Probabilistic Reasoning:} Recognizing that high-value domains are inherently stochastic, evaluation frameworks should evolve to reward \textbf{calibration over false confidence}. Future agents should be developed to output probabilistic forecasts (e.g., confidence intervals, top-k distributions). This shift prioritizes the agent's ability to model risk and volatility, which is far more valuable than generating a single, potentially hallucinatory number.
    
    \item \textbf{Disentangling Capability from Safety:} To improve utility in sensitive sectors like Public Health, alignment strategies must distinguish between avoiding harm (Policy-Based Refusal) and failing to find data (Capability-Based Refusal). A truly capable agent for high-value domains must be bold enough to retrieve objective safety data while remaining conservative enough to avoid misinformation.
\end{itemize}

In summary, FutureX-Pro serves as a rigorous stress test for the next generation of LLMs. It underscores that mastering high-value domains requires more than scaling parameters; it requires the precise integration of specialized search, quantitative reasoning, and deep domain grounding.
\newpage
\section{Contributions}

\textbf{Project Lead:} Jiashuo Liu, Siyuan Chen, Zaiyuan Wang, Zhiyuan Zeng

\textbf{Core Contributors:} (by domain)
\begin{itemize}
    \item FutureX-Finance: Jiacheng Guo$^\dagger$, Liang Hu, Lingyue Yin, Suozhi Huang$^\dagger$, Wenxin Hao, Yang Yang, Zerui Cheng, Zixin Yao$^\dagger$
    \item FutureX-Retail: Lingyue Yin, Wenxin Hao
    \item FutureX-PublicHealth: Haoxin Liu$^\star$, Jiayi Cheng, Yuzhen Li, Zezhong Ma
    \item FutureX-NaturalDisaster: Haoxin Liu$^\star$
    \item FutureX-Search: Bingjie Wang, Bingsen Qiu, Xiao Liu, Zeyang Zhang, Zijian Liu
    \item Infrastructure: Jinpeng Wang, Mingren Yin, Tianci He, Yali Liao, Yixiao Tian, Zhenwei Zhu
\end{itemize}

\textbf{Contributors:}
Anqi Dai, Ge Zhang, Jingkai Liu, Kaiyuan Zhang, Wenlong Wu,  Xiang Gao, Xinjie Chen, Zhixin Yao, Zhoufutu Wen

\textbf{Advisors:}
B. Aditya Prakash$^\star$, Jose Blanchet$^\ddagger$, Mengdi Wang$^\dagger$, Nian Si$^\gamma$, Wenhao Huang

$^\dagger$ Princeton University\\
$^\star$ Georgia Institute of Technology\\
$^\ddagger$ Stanford University\\
$^\gamma$ Hong Kong University of Science and Technology

Authors are listed alphabetically by first name. Unless specified, all authors are from ByteDance Seed team.

\clearpage

\bibliographystyle{plainnat}
\bibliography{main}

@article{zeng2025futurex,
  title={Futurex: An advanced live benchmark for llm agents in future prediction},
  author={Zeng, Zhiyuan and Liu, Jiashuo and Chen, Siyuan and He, Tianci and Liao, Yali and Tian, Yixiao and Wang, Jinpeng and Wang, Zaiyuan and Yang, Yang and Yin, Lingyue and others},
  journal={arXiv preprint arXiv:2508.11987},
  year={2025}
}

@inproceedings{chen2021finqa,
  title={Finqa: A dataset of numerical reasoning over financial data},
  author={Chen, Zhiyu and Chen, Wenhu and Smiley, Charese and Shah, Sameena and Borova, Iana and Langdon, Dylan and Moussa, Reema and Beane, Matt and Huang, Ting-Hao and Routledge, Bryan R and others},
  booktitle={Proceedings of the 2021 Conference on Empirical Methods in Natural Language Processing},
  pages={3697--3711},
  year={2021}
}

@article{jin2020disease,
  title={What Disease does this Patient Have? A Large-scale Open Domain Question Answering Dataset from Medical Exams},
  author={Jin, Di and Pan, Eileen and Oufattole, Nassim and Weng, Wei-Hung and Fang, Hanyi and Szolovits, Peter},
  journal={arXiv preprint arXiv:2009.13081},
  year={2020}
}

@inproceedings{zhou2024webarena,
  title={WEBARENA: A REALISTIC WEB ENVIRONMENT FOR BUILDING AUTONOMOUS AGENTS},
  author={Zhou, Shuyan and Xu, Frank F and Zhu, Hao and Zhou, Xuhui and Lo, Robert and Sridhar, Abishek and Cheng, Xianyi and Ou, Tianyue and Bisk, Yonatan and Fried, Daniel and others},
  booktitle={12th International Conference on Learning Representations, ICLR 2024},
  year={2024}
}

@article{white2024livebench,
  title={Livebench: A challenging, contamination-free llm benchmark},
  author={White, Colin and Dooley, Samuel and Roberts, Manley and Pal, Arka and Feuer, Ben and Jain, Siddhartha and Shwartz-Ziv, Ravid and Jain, Neel and Saifullah, Khalid and Naidu, Siddartha and others},
  journal={arXiv preprint arXiv:2406.19314},
  volume={4},
  year={2024}
}

@inproceedings{kargerforecastbench,
  title={ForecastBench: A Dynamic Benchmark of AI Forecasting Capabilities},
  author={Karger, Ezra and Bastani, Houtan and Yueh-Han, Chen and Jacobs, Zachary and Halawi, Danny and Zhang, Fred and Tetlock, Philip},
  booktitle={The Thirteenth International Conference on Learning Representations}
}

@article{deng2023mind2web,
  title={Mind2web: Towards a generalist agent for the web},
  author={Deng, Xiang and Gu, Yu and Zheng, Boyuan and Chen, Shijie and Stevens, Sam and Wang, Boshi and Sun, Huan and Su, Yu},
  journal={Advances in Neural Information Processing Systems},
  volume={36},
  pages={28091--28114},
  year={2023}
}

@inproceedings{mialon2023gaia,
  title={Gaia: a benchmark for general ai assistants},
  author={Mialon, Gr{\'e}goire and Fourrier, Cl{\'e}mentine and Wolf, Thomas and LeCun, Yann and Scialom, Thomas},
  booktitle={The Twelfth International Conference on Learning Representations},
  year={2023}
}

@inproceedings{liu2024agentbench,
  title={AgentBench: Evaluating LLMs as Agents},
  author={Liu, Xiao and Yu, Hao and Zhang, Hanchen and Xu, Yifan and Lei, Xuanyu and Lai, Hanyu and Gu, Yu and Ding, Hangliang and Men, Kaiwen and Yang, Kejuan and others},
  booktitle={ICLR},
  year={2024}
}

@incollection{garrett2017challenge,
  title={The challenge of global health},
  author={Garrett, Laurie},
  booktitle={Global Health},
  pages={525--548},
  year={2017},
  publisher={Routledge}
}

@article{wei2025browsecomp,
  title={Browsecomp: A simple yet challenging benchmark for browsing agents},
  author={Wei, Jason and Sun, Zhiqing and Papay, Spencer and McKinney, Scott and Han, Jeffrey and Fulford, Isa and Chung, Hyung Won and Passos, Alex Tachard and Fedus, William and Glaese, Amelia},
  journal={arXiv preprint arXiv:2504.12516},
  year={2025}
}

@article{ritchie2022natural,
  title={Natural disasters},
  author={Ritchie, Hannah and Rosado, Pablo and Roser, Max},
  journal={Our world in data},
  year={2022}
}

@article{albreht2023challenges,
  title={Challenges to global health emerging from the COVID-19 pandemic},
  author={Albreht, Tit},
  journal={Sustainability},
  volume={15},
  number={9},
  pages={7633},
  year={2023},
  publisher={MDPI}
}

@article{yu2018big,
  title={Big data in natural disaster management: a review},
  author={Yu, Manzhu and Yang, Chaowei and Li, Yun},
  journal={Geosciences},
  volume={8},
  number={5},
  pages={165},
  year={2018},
  publisher={MDPI}
}

@article{zhu2025findeepresearch,
  title={FinDeepResearch: Evaluating Deep Research Agents in Rigorous Financial Analysis},
  author={Zhu, Fengbin and Ng, Xiang Yao and Liu, Ziyang and Liu, Chang and Zeng, Xianwei and Wang, Chao and Tan, Tianhui and Yao, Xuan and Shao, Pengyang and Xu, Min and others},
  journal={arXiv preprint arXiv:2510.13936},
  year={2025}
}

@article{dong2025finch,
  title={Finch: Benchmarking Finance \& Accounting across Spreadsheet-Centric Enterprise Workflows},
  author={Dong, Haoyu and Zhang, Pengkun and Gao, Yan and Dong, Xuanyu and Cheng, Yilin and Lu, Mingzhe and Yakefu, Adina and Zheng, Shuxin},
  journal={arXiv preprint arXiv:2512.13168},
  year={2025}
}

@article{min2025ecombench,
  title={EcomBench: Towards Holistic Evaluation of Foundation Agents in E-commerce},
  author={Min, Rui and Qiao, Zile and Xu, Ze and Zhai, Jiawen and Gao, Wenyu and Chen, Xuanzhong and Sun, Haozhen and Zhang, Zhen and Wang, Xinyu and Zhou, Hong and others},
  journal={arXiv preprint arXiv:2512.08868},
  year={2025}
}

@article{guo2025cryptobench,
  title={CryptoBench: A Dynamic Benchmark for Expert-Level Evaluation of LLM Agents in Cryptocurrency},
  author={Guo, Jiacheng and Huang, Suozhi and Yao, Zixin and Zhang, Yifan and Lu, Yifu and Liu, Jiashuo and Li, Zihao and Deng, Yanyan and Xiao, Qixin and Tian, Jia and others},
  journal={arXiv preprint arXiv:2512.00417},
  year={2025}
}

@article{qiu2025bmgq,
  title={BMGQ: A Bottom-up Method for Generating Complex Multi-hop Reasoning Questions from Semi-structured Data},
  author={Qiu, Bingsen and Liu, Zijian and Liu, Xiao and Wang, Bingjie and Zhang, Feier and Qin, Yixuan and Li, Chunyan and Yang, Haoshen and Gao, Zeren},
  journal={arXiv preprint arXiv:2510.24151},
  year={2025}
}

@article{hu2025finsearchcomp,
  title={Finsearchcomp: Towards a realistic, expert-level evaluation of financial search and reasoning},
  author={Hu, Liang and Jiao, Jianpeng and Liu, Jiashuo and Ren, Yanle and Wen, Zhoufutu and Zhang, Kaiyuan and Zhang, Xuanliang and Gao, Xiang and He, Tianci and Hu, Fei and others},
  journal={arXiv preprint arXiv:2509.13160},
  year={2025}
}

\end{document}